\documentclass[
]{ceurart}

\usepackage{subcaption}
\usepackage{siunitx}
\usepackage{listings}
\usepackage{soul}

\renewcommand{\bfseries}{\fontseries{b}\selectfont}
\newrobustcmd{\B}{\bfseries}

\begin{document}

\copyrightyear{2023}
\copyrightclause{Copyright for this paper by its authors.
  Use permitted under Creative Commons License Attribution 4.0
  International (CC BY 4.0).}

\conference{CLEF 2023: Conference and Labs of the Evaluation Forum, September 18–21, 2023, Thessaloniki, Greece}

\title{AI-UPV at EXIST 2023 -- Sexism Characterization Using Large Language Models Under\\ The Learning with Disagreements Regime}

\title[mode=sub]{Notebook for the EXIST Lab at CLEF 2023}


\address[1]{Universitat Politècnica de València,  Valencia, Spain}
\address[2]{RMIT University, Melbourne, Australia}
\address[3]{University of Milano-Bicocca,  Milan, Italy}

\author[1,2]{Angel Felipe {Magnossão de Paula}}[%
orcid=0000-0001-8575-5012,
email=adepau@doctor.upv.es,
]

\author[1,3]{Giulia Rizzi}[%
orcid=0000-0002-0619-0760,
email=g.rizzi10@campus.unimib.it,
]
\cormark[1]

\author[3]{Elisabetta Fersini}[%
orcid=0000-0002-8987-100X,
email=elisabetta.fersini@unimib.it,
]

\author[2]{Damiano Spina}[
orcid=0000-0001-9913-433X,
email=damiano.spina@rmit.edu.au,
]

\cortext[1]{Corresponding author.}

\begin{abstract}
With the increasing influence of social media platforms, it has become crucial to develop automated systems capable of detecting instances of sexism and other disrespectful and hateful behaviors to promote a more inclusive and respectful online environment.
Nevertheless, these tasks are considerably challenging considering different hate categories and the author's intentions, especially under the learning with disagreements regime. 
This paper describes AI-UPV team's participation in the EXIST (sEXism Identification in Social neTworks) Lab at CLEF 2023~\cite{plaza2023overview,plaza2023extended}.
%
The proposed approach aims at addressing the task of sexism identification and characterization under the learning with disagreements paradigm by training directly from the data with disagreements, without using any aggregated label. Yet, performances considering both soft and hard evaluations are reported.
The proposed system uses large language models (i.e., mBERT and XLM-RoBERTa) and ensemble strategies for sexism identification and classification in English and Spanish.
In particular, our system is articulated in three different pipelines. 
The ensemble approach outperformed the individual large language models obtaining the best performances both adopting a soft and a hard label evaluation.
This work describes the participation in all the three EXIST tasks, considering a soft evaluation, it obtained fourth place in Task 2 at EXIST and first place in Task 3, with the highest ICM-Soft of $-2.32$ and a normalized ICM-Soft of $0.79$. The source code of our approaches is publicly available at \url{https://github.com/AngelFelipeMP/Sexism-LLM-Learning-With-Disagreement}.


\end{abstract}


\maketitle

\section{Introduction}
Sexism in online content represents a significant challenge in maintaining inclusive and fair digital environments.
In addition, cultural differences \cite{schlicht2021upv}, misinformation \cite{Ipek2021TREC}, and hateful behavior \cite{schlicht2021unified} aggravate it.
Many women from all around the world have reported abuses, discrimination, and other sexist situations in real life on social media. Social networks are also contributing to the spread of sexism and other disrespectful and hateful behaviors. While social platforms such as Twitter are constantly developing new ways to identify and eliminate hateful content, they face numerous challenges when dealing with the massive amount of data generated by users. In this context, automated tools are adopted not only to assist in detecting and alerting against sexist behaviors, but also in estimating how frequently sexist and abusive phenomena are faced on social media platforms, what forms of sexism are more diffused, and how those harmful messages are spread.\\
Traditional sexism detection systems often rely on predefined labels and fixed perspectives, which may fail to capture the complexity and subjectivity inherent in sexist statements. Detecting and mitigating sexism remains a challenging task due to the subjective nature of its assessment.
Perspectivism, on the other hand, provides a viable approach for improving detection by incorporating multiple opinions and viewpoints.\\
An important contribution in this field, that aims at tackling the problem of sexism identification under the paradigm of learning with disagreements has been given by the \textit{EXIST 2023: sEXism Identification in Social neTworks} 
 \cite{plaza2023overview,plaza2023extended}.\\
In this paper, we address all the tasks proposed in the challenge, which cover different granularities and perspectives of sexism classification. In particular, we proposed a Large Language Model (LLM) ensemble-based strategy to predict both soft levels that reflect the set of labels provided by the annotators and hard labels derived from the aggregation of different annotators' perspectives.\\
The paper is organized as follows. An overview of the state of the art is provided in Section \ref{sec:soa} focusing both on Sexism Detection and Perspectivism. Details about the shared task and the related datasets are reported in Section \ref{sec:task} and Section \ref{sec:dataset} respectively. In Section \ref{sec:approach} the proposed approach is detailed focusing on the adopted prediction models, the proposed ensemble approach, and the post-processing operations. In Section \ref{sec:results}, the results achieved by the proposed models are reported. 
Finally, conclusions and future research directions are summarized in Section \ref{sec:conclusion}.

\section{Related Work}\label{sec:soa}

Sexism, defined as prejudice, stereotyping, or discrimination based on gender, is a pervasive issue affecting individuals across various contexts. The advent of digital platforms has amplified the reach and impact of sexist content, necessitating the development of automated systems to detect and counteract sexism. Researchers have proposed several approaches to address this issue, ranging from rule-based methods \cite{samory2021call}  to more advanced machine-learning techniques to  address the problem of sexism detection mostly from a linguistic perspective \cite{de2021sexism, de2022detection}, with only a few attempts to tackle the problem from a visual or multimodal point of view \cite{campisi2018automatic, fersini2019detecting}. \\
The majority of state-of-the-art approaches leveraged machine learning techniques, particularly relying on pre-trained LLM. \\
However, traditional approaches to sexism detection often overlook the complexity and subjectivity inherent in describing sexist behavior. \\
Sexism can in fact be further categorized in different forms according to the intention of the author, or the type of sexism. 
In the EXIST 2021 and 2022  challenges \cite{rodriguez2021overview, rodriguez2022overview} five classes of sexism have been detected: ``Ideological and inequality'', ``Stereotyping and dominance'', ``Objectification'', ``Sexual violence'' and ``Misogyny and non-sexual violence''. 
Similarly at SemEval 2023 -- Task 10 -- Explainable Detection of Online Sexism (EDOS) \cite{kirk2023semeval}, provided a new taxonomy for the more explainable classification of sexism in three hierarchical levels – binary sexism detection, category of sexism, and fine-grained vector of sexism.
As emerged from the leaderboard, top systems use multiple models or ensembles, and many top systems apply further pre-training and multi-task learning \cite{de2021sexism, rydelek2023adamr, rallabandi2023sss}.\\
Despite the recent advances in employing attention mechanisms and other deep learning approaches, the detection of sexism \cite{de2022detection} and hate speech \cite{magnossao2021ai} in general is still considered a major challenge, especially, when dealing with social media text written in low-resource languages, such as Arabic \cite{de2022upv}.\\
Multilingual LLM, especially if encapsulated in Ensemble methods, have been proven to be a robust solution to address this task \cite{depaula2023transformers, de2021sexism}.
A considerable contribution in the field is represented by challenges' tasks like AMI IberEval, Evalita and EXIST.
AMI IberEval 2018 \cite{fersini2018overview} proposed two shared tasks mainly focused on tackling the problem of misogyny on Twitter, in three different languages, namely English, Italian, and Spanish.\\
Despite the problem of sexism is mostly addressed from a textual perspective, a growing field of research is tackling the problem from visual or multimodal point of view. Sexist content can in fact be represented also in images or in a multimodal form. 
From a visual perspective, most of the available investigations relate to offensive, non-compliant, or pornographic content detection \cite{gandhi2020scalable, tabone2021pornographic, SoAPornographyDetection}, while from a multimodal point of view, the main contribution is given by the \textit{Hateful Meme challenge: Detecting Hate Speech in multimodal Memes} \cite{kiela2020hateful}, where one of the targets of hateful memes were women and by \textit{SemEval-2022 Task 5: Multimedia Automatic Misogyny Identification (MAMI)}\cite{fersini2022semeval} that focused on misogyny recognition in memes.\\
Finally, many research papers give emphasis to an open issue in sexism detection: the presence of bias that could affect the real performance of the models. In this research area, bias analysis is still in its infancy with only a few contributions addressing the problem considering textual data and sexism identification \cite{alba2022bias, wiegand2019detection, zhou2021challenges, campisi2018automatic}.\\
While the aforementioned studies have made substantial contributions to detecting sexism, they frequently approach the problem as an objective concept without considering taking into account the complexities and differing perspectives around it. Sexism may, in fact, take several forms depending on cultural, societal, and individual factors.\\
Recognizing the importance of addressing \emph{perspectivism}, some researchers have investigated methods to include a more nuanced understanding of the tasks in detection systems \cite{uma2021learning}.  
However, only a few of them investigated perspectivism in sexism detection. For example, 
\citet{kalra2021sexism} explore the errors made by classification models and discusses the difficulty in automatically classifying sexism due to the subjectivity of the labels and the complexity of natural language used in social media.\\
The attention to perspectivism is also reflected in the challenges with an increasing number of tasks that consider the level of agreement of the annotators in the form of soft labels.
For instance, EXIST 2023 \cite{plaza2023overview,plaza2023extended}, addresses sexism identification at different granularities and perspectives under the learning with disagreements paradigm.


\section{Task Description}\label{sec:task}

EXIST aims to detect sexism in a broad sense, from explicit misogyny to other subtle expressions that involve implicit sexist behaviors (EXIST 2021 \cite{rodriguez2021overview}, EXIST 2022 \cite{rodriguez2022overview}).  
The main task of the new edition of EXIST is to develop models able to capture sexism in all its forms, 
while considering the perspective of the learning with disagreements paradigm.\\
The EXIST Lab at CLEF 2023 \cite{plaza2023overview,plaza2023extended} is articulated in three different tasks that address sexism identification at different granularities and perspectives:

\paragraph{TASK 1: Sexism Identification.}
The first task addresses sexism identification as a binary problem, requiring the systems to classify if a tweet contains sexist expressions or behaviors (i.e., it is sexist itself, describes a sexist situation or criticizes a sexist behavior) or not.

\paragraph{TASK 2: Source Intention.}
The second task focuses on classification of the message in accordance with the author's intentions (distinguishing between ``Direct'', ``Reported'', and ``Judgemental''), which sheds light on the part social networks play in the creation and spread of sexist messages. 

\paragraph{TASK 3: Sexism Categorization.}
The third task aims to categorize tweet into different types of sexism (i.e., ``Ideological and Inequality'', ``Stereotyping and Dominance'', ``Objectification'', ``Sexual Violence'', and ``Misogyny and Non-sexual Violence'') that reflect different focus of sexism attitudes ranging from domestic and parenting roles, career opportunities to sexual image.

\noindent The challenge also faces the sexism identification from the perspective of the learning with disagreements paradigm introducing two evaluations approaches for each task:
\begin{itemize}
    \item \textbf{Soft Evaluation}. This is the evaluation under the learning with disagreements paradigm. Systems performances are evaluated through a soft-soft evaluation that compares the probabilities assigned by the system with the probabilities assigned by the set of human annotators. In this case, the evaluation measure is ICM-Soft~\cite{plaza2023overview, plaza2023extended}, while  additional evaluation measures such as Cross Entropy are also reported.
    \item \textbf{Hard Evaluation}. Systems performances are also evaluated on the hard label derived from the different annotators’ labels, through a probabilistic threshold computed for each task. The adopted evaluation measure is the original ICM \cite{amigo2022evaluating}. F1 Score is also reported.
    
\end{itemize}
Additionally, for systems that provide a hard output a hard-soft evaluation is evaluated, comparing the categories assigned by the system with the probabilities assigned to each category in the ground truth with ICM-Soft. 

\section{Dataset}\label{sec:dataset}
The EXIST 2023 dataset incorporates multiple types of sexist expressions, including descriptive or reported assertions where the sexist message is a description of sexist behavior.\\
In particular, the dataset is composed of more than 10,000 tweets both in English and Spanish, divided into a test set (2,076 tweets), a development set (1,038 tweets), and a training set (6,920 tweets). 

For each sample, the following attributes are provided in a JSON format:
\begin{itemize}
    \item \textbf{id\_EXIST}: a unique identifier for the tweet.
    \item \textbf{lang}: the language of the text (``en'' or ``es'').
    \item \textbf{tweet}: the text of the tweet.
\item \textbf{number\_annotators}: the number of persons that have annotated the tweet.
\item \textbf{annotators}: a unique identifier for each of the annotators.
\item \textbf{gender\_annotators}: the gender of the different annotators (``F'' or ``M'', for female and male respectively).
\item \textbf{age\_annotators}: the age group of the different annotators (grouped in ``18--22'', ``23--45'', or ``46+'').
\item \textbf{labels\_task1}: a set of labels (one for each of the annotators) that indicate if the tweet
contains sexist expressions or refers to sexist behaviors or not (``YES'' or ``NO'').
\item \textbf{labels\_task2}: a set of labels (one for each of the annotators) recording the intention
of the person who wrote the tweet (``DIRECT'', ``REPORTED'', ``JUDGEMENTAL'', ``--'', and ``UNKNOWN'').
\item \textbf{labels\_task3}: a set of arrays of labels (one array for each of the annotators) indicating the type or types of sexism that are found in the tweet (``IDEOLOGICAL-INEQUALITY'', ``STEREOTYPING-DOMINANCE'', ``OBJECTIFICATION'', ``SEXUAL-VIOLENCE'', ``MISOGYNY-NON-SEXUAL-VIOLENCE'', ``--'', and ``UNKNOWN'').
\item \textbf{split}: subset within the dataset the tweet belongs to (``TRAIN'', ``DEV'', ``TEST'' + ``EN''/``ES'').
\end{itemize}
It is important to note that since hard labels for Tasks 2 and 3 are assigned only if the tweet has been labeled as sexist (label ``YES'' for Task 1), the label ``--'' is assigned to not sexist tweets in Tasks 2 and 3, while the label ``UNKNOWN'' is assigned to those tweets where the annotators did not provide a label.
The test set includes only the following attributes: ``id\_EXIST'', ``lang'', ``tweet'', and ``split''.


\section{Proposed Approach}\label{sec:approach}
In this Section, the proposed approach to address all the tasks is described.\\
Inspired by the state-of-the-art results, we propose an ensemble method to combine two transformer-based models: namely mBERT and XLM-RoBERTa. The proposed approach focuses on soft label predictions, while the hard labels are directly derived by selecting the most probable ones.



\paragraph{BERT Base Multilingual.}
BERT Base Multilingual \cite{devlin2019bert}, also called ``mBERT'', is a widely used language model that has made substantial progress in the field of natural language processing. mBERT performs well in tasks like text classification, named entity identification, and question answering thanks to its bidirectional context. Additionally, mBERT has been pre-trained on a large corpus of text from various languages, enabling it to capture language-specific patterns and nuances. \\
mBERT has been used by researchers to successfully identify and categorize hate speech across a variety of areas and languages \cite{schutz2021automatic, bengoetxea2022multiaztertest}. Furthermore, mBERT is an effective tool for identifying hate speech and creating safer online settings because of its contextual awareness and capacity to identify subtle linguistic patterns.

\paragraph{XLM-RoBERTa.}
Cross-lingual Language Model-RoBERTa \cite{conneau2019unsupervised}, also called ``XLM-RoBERTa'', is a state-of-the-art language model that combines the benefits of cross-lingual pretraining and RoBERTa. 
The usage of parallel data from various languages during pre-training enables it to effectively transfer information between languages and  therefore to comprehend and produce text in a variety of languages. XLM-RoBERTa was adopted on a variety of natural language processing tasks, including text categorization, named entity identification, and machine translation. XLM-RoBERTa is widely adopted by researchers in multilingual situations because of its robustness in managing multilingual data and its proficiency in cross-lingual transfer.\\
Moreover, researchers have used XLM-RoBERTa to identify hate speech in multilingual settings, a task for which the model's capacity to handle several languages is crucial. It has been demonstrated that XLM-RoBERTa can detect hate speech in a variety of languages, including English, Spanish, and Arabic \cite{ali2022hate, rezaul2022multimodal, roy2022hate}. It can generalize well and identify hate speech in many language situations thanks to its cross-linguistic transfer abilities.\\


\noindent The proposed approach is articulated in the following steps that have been repeated for each EXIST task:
\begin{itemize}
    \item \textbf{Hyperparameters Selection.}
    The aim of the first phase is to identify the most suitable number of fine-tuning epochs for our two LLMs. The two selected models have been implemented using the hugging face framework \cite{huggingface}  and fine-tuned in the EXIST training dataset with 1 up to 10 epochs. Each model has then been evaluated on the EXIST validation dataset.
\item \textbf{Model Training.}
The training phase has the main goal of learning directly with soft labels.
The parameter identified through the previous phase has been used to fine-tune the final models, on the whole dataset obtained via the union of the training and validation dataset.
\item \textbf{Model Predictions.} 
The predictions related to the soft labels, both from the fine-tuned single models and from their aggregation given by an ensemble model, have been computed. 
First, we obtain the predicted probabilities for each model separately on the EXIST test set. Then, we combined the probabilities given by the mBERT and XLM-RoBERTa models by estimating the mean of their predicted probabilities. 
\item  \textbf{Prediction Adjustment.} To ensure that the predicted probabilities are compliant with respect to the number of annotators, we performed the following operation: given the probability distribution of a model, we selected the most similar distribution according to cosine similarity with the feasible distributions.
This operation allowed adjusting the obtained discrete probability distribution to the nearest ones that match the number of annotators, while ensuring a stochastic distribution for Tasks 1 and 2. This ensures that the final prediction is consistent with the number of annotators and avoids any potential errors that may arise from the usage of raw predictions.

\end{itemize}
The hard labels can be derived directly from the predicted soft labels by selecting the most probable label. The proposed framework has been instantiated according to the following three different approaches, depicted in Figure \ref{fig:predictions}:

\begin{figure}[h]
    \centering
    \includegraphics[width =\textwidth]{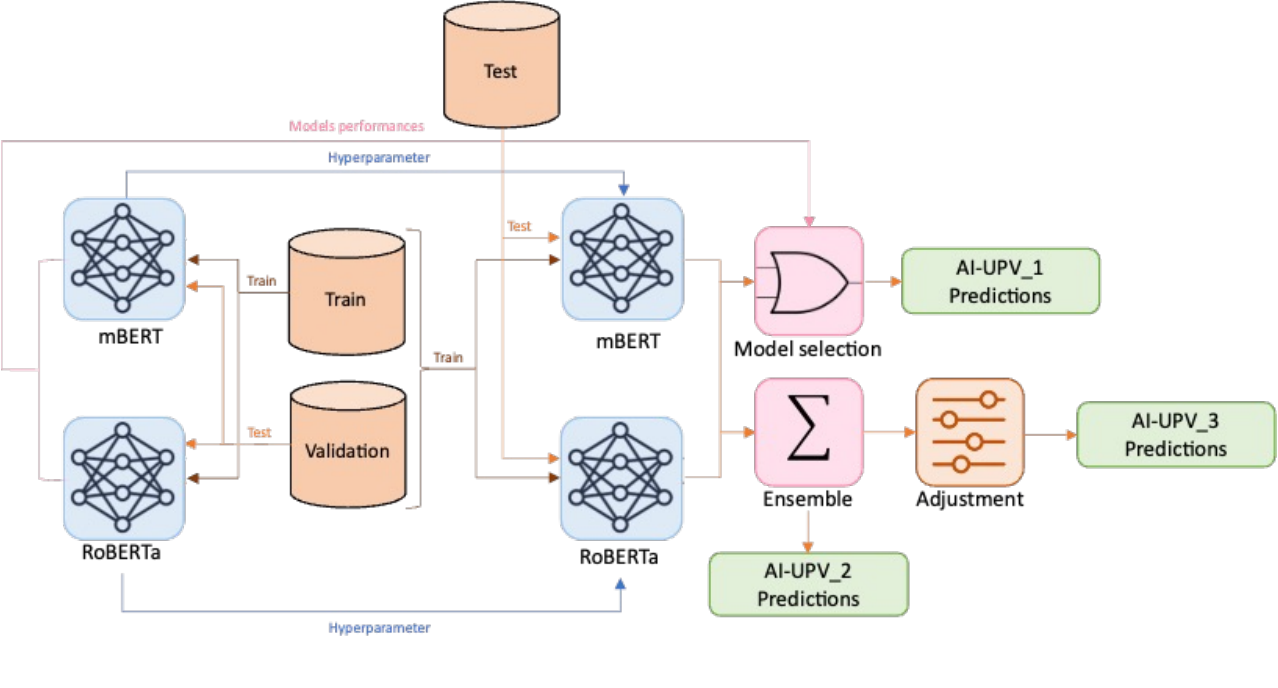}
    \caption{Schematic representation of the proposed pipeline.}
    \label{fig:predictions}
\end{figure}


\paragraph{AI-UPV\_1: Model Selection.}
The first submission refers to the predictions returned by the best-performing model. In particular, both mBERT and XLM-RoBERTa are trained on EXIST training data and evaluated on the EXIST validation data in order to identify, not only the most suitable number of fine-tuning epochs (as described above) but also to identify the best performing model for each of the EXIST tasks. The predictions obtained from the selected model (i.e., BERT or XLM-RoBERTa) trained on the whole EXIST dataset (obtained by the union of the training and validation datasets) are reported as AI-UPV\_1.

\paragraph{AI-UPV\_2: Ensemble Model.}
The second submission refers to the predictions returned by a naive ensemble model. In particular, the final label is determined through a mean aggregation function that considers both fine-tuned models 
(i.e., mBERT and XLM-RoBERTa).

\paragraph{AI-UPV\_3: Adjusted Ensemble Model.}
The third submission refers to the adjusted  ensemble model predictions. Once the hyperparameters have been selected and the models have been trained, the ensemble computes the average probabilities considering the predictions given by mBERT and XLM-RoBERTa to then adjust such estimations according to the number of evaluators.


\section{Results and Discussion}\label{sec:results}
In this section, the results obtained through our participation in the challenge are reported and compared with the baselines provided by the organizers. Regarding the \textbf{Baselines}, three main approaches have been considered:
\begin{itemize}
    \item \textbf{Majority Class}: non-informative baseline that classifies all instances according to the majority class.
    \item \textbf{Minority Class}: non-informative baseline that classifies all instances using the minority class.
     \item \textbf{Gold}: since the ICM measure is unbounded, a baseline based on an oracle that perfectly predicts the ground truth is considered to provide the best possible reference.
    
\end{itemize}
We report in the following tables the results obtained by the proposed approaches and the baselines, distinguishing between the tasks and the languages considered in the challenge.  Table \ref{tab:res-task1-ALL} shows the results for EXIST Task 1 on all (English and Spanish) instances; the best-performing model (i.e., AI-UPV\_2) achieved seventh place in the global ranking considering a soft evaluation.

\begin{table}[h]
\caption{Results of the proposed approaches for Task 1 on all test instances (English + Spanish).}
\label{tab:res-task1-ALL}
\resizebox{\textwidth}{!}{%
\sisetup{detect-weight=true,mode=text,
  round-mode = places,      
  round-precision = 2,      
}
\begin{tabular}{lSSSSSS}
\toprule
                            & {ICM-Soft} & {ICM-Soft   Norm} & {Cross   Entropy} & {ICM-Hard} & {ICM-Hard   Norm} & {F1}     \\
\midrule
EXIST2023\_gold & 3.1182   & 1             & 0.5472          & 0.9948   & 1               & 1      \\
EXIST2023\_majority\_class &	-2.3585	 &0.1152	&4.6115 & -0.4413&	0.0847&	0\\
EXIST2023\_minority\_class &	-3.0717&	0&	5.3572& -0.5742 &	0&	0.5698\\
\midrule
AI-UPV\_1  (Model Selection)                 & 0.5448   & 0.5843          & 1.5543          & 0.47     & 0.6655          & 0.7445 \\
AI-UPV\_2  (Ensemble)                 &  \B 0.7343   &  \B 0.6149          &\B 1.3607          & \B 0.5106   & \B 0.6914          & \B 0.7589 \\
AI-UPV\_3  (Adj. Ensemble)                 & 0.6772   & \B 0.6056          & 1.64            & \B 0.5119  & \B 0.6922          & \B 0.7574 \\
\bottomrule
\end{tabular}%
}
\end{table}

\noindent Analogously Tables \ref{tab:res-task1-ES} and \ref{tab:res-task1-EN} present the results for  Task 1 on Spanish and English instances, respectively. In this case, we can highlight that the proposed approaches perform in an analogous way in both languages, considering that no strong performance variations between them can be observed.

\begin{table}[h]
\caption{Results of the proposed approaches for Task 1 on ES (Spanish) instances.}
\label{tab:res-task1-ES}
\resizebox{\textwidth}{!}{%
\sisetup{detect-weight=true,mode=text,
  round-mode = places,      
  round-precision = 2,      
}
\begin{tabular}{lSSSSSS}
\toprule
                            & {ICM-Soft} & {ICM-Soft   Norm} & {Cross   Entropy} & {ICM-Hard} & {ICM-Hard   Norm} & {F1}     \\
\midrule
EXIST2023\_gold & 3.1177   & 1               & 0.5208          & 0.9999   & 1               & 1      \\
EXIST2023\_majority\_class &	-2.5421	&0.0056&	4.9631 &-0.4897&	0.0138&	0\\
EXIST2023\_minority\_class&	-2.5742	&0&	5.0055&-0.5106&	0&	0.6066\\
\midrule
AI-UPV\_1  (Model Selection)   & 0.5956   & 0.5569          & 1.5177          & 0.4553   & 0.6395          & 0.7644 \\
AI-UPV\_2  (Ensemble)                   & \B 0.7415   & \B 0.5825          & \B 1.3259          & 0.489    & 0.6618          & 0.7731 \\
AI-UPV\_3  (Adj. Ensemble)                 & 0.6999   & \B 0.5752          & 1.6134          & \B 0.5069   & \B 0.6736          & \B 0.7753 \\
\bottomrule
\end{tabular}%
}
\end{table}

\begin{table}[h]
\caption{Results of the proposed approaches for Task 1 on EN (English) instances.}
\label{tab:res-task1-EN}
\resizebox{\textwidth}{!}{%
\sisetup{detect-weight=true,mode=text,
  round-mode = places,      
  round-precision = 2,      
}
\begin{tabular}{lSSSSSS}
                            \toprule
                            & {ICM-Soft} & {ICM-Soft   Norm} & {Cross   Entropy} & {ICM-Hard} & {ICM-Hard   Norm} & {F1}     \\
\midrule
EXIST2023\_gold & 3.1141   & 1               & 0.577           & 0.9798   & 1               & 1      \\
EXIST2023\_majority\_class&	-2.1991&	0.2333	&4.2166 & -0.3965&	0.163	&0\\
EXIST2023\_minority\_class&	-3.8158	&0&	5.7521 &-0.6646&	0&	0.526\\
\midrule
AI-UPV\_1  (Model Selection)                      & 0.4204   & 0.6113          & 1.5954          & 0.4705   & 0.6903          & 0.7176 \\
AI-UPV\_2  (Ensemble)                   & \B 0.67     &\B 0.6473          & \B 1.3998          & \B0.5211   & \B 0.7211          & \B 0.7399 \\
AI-UPV\_3  (Adj. Ensemble)                   & 0.5936   & 0.6363          & 1.67            & 0.5041   & 0.7107          & 0.7334\\
\bottomrule
\end{tabular}%
}

\end{table}
\noindent Considering the differences of the ICM-Soft -- which is the official evaluation measure -- it is possible to conclude that: (i) all the proposed models outperformed the proposed baselines, both considering all samples and the language-dependent splits, (ii) the inclusion of an ensemble approach introduces a significant improvement in performances, and (iii) the adjustment operation slightly penalizes the ensemble performances. \\
We report in  Tables \ref{tab:res-task2-all}, \ref{tab:res-task2-ES} and \ref{tab:res-task2-EN},  the results related to the Exist Task 2, distinguishing between the results on the overall dataset and  single languages. Regarding the overall dataset, the proposed AI-UPV\_2 achieved the fourth place in the global ranking considering the soft evaluation. 


\begin{table}[h]
\caption{Results of the proposed approaches for Task 2 on all test instances (English + Spanish). }
\label{tab:res-task2-all}
\resizebox{\textwidth}{!}{%
\sisetup{detect-weight=true,mode=text,
  round-mode = places,      
  round-precision = 2,      
}
\begin{tabular}{lSSSSSS}
\toprule
                            & {ICM-Soft} & {ICM-Soft   Norm} & {Cross   Entropy} & {ICM-Hard} & {ICM-Hard   Norm} & {F1}     \\
                               \midrule
                               
EXIST2023\_gold    & 6.2057   & 1               & 0.9128          & 1.5378   & 1               & 1      \\
EXIST2023\_majority\_class & -5.446   & 0.7025          & 4.6233          & -0.9504  & 0.4697          & 0.1603 \\
EXIST2023\_minority\_class & -32.9552 & 0               & 8.8517          & -3.1545  & 0               & 0.028  \\
\midrule
AI-UPV\_1  (Model Selection)                    & -4.3632  & 0.7301          & 3.0172          & 0.1217   & 0.6982          & 0.4897 \\
AI-UPV\_2  (Ensemble)                      & \B -1.6836  & \B 0.7985          & \B 2.1697          & \B 0.1951   & \B 0.7139          & \B 0.4962 \\
AI-UPV\_3  (Adj. Ensemble)                 & -1.7691  & \B 0.7964          & 2.5424          & 0.187    & \B 0.7121          & \B 0.4993 \\
        \bottomrule\\
\end{tabular}%
}
\end{table}

\begin{table}[!h]
\caption{Results of the proposed approaches for Task 2 on ES (Spanish) instances}
\label{tab:res-task2-ES}
\resizebox{\textwidth}{!}{%
\sisetup{detect-weight=true,mode=text,
  round-mode = places,      
  round-precision = 2,      
}
\begin{tabular}{lSSSSSS}
                               \toprule
                               & {ICM-Soft} & {ICM-Soft   Norm} & {Cross   Entropy} & {ICM-Hard} & {ICM-Hard   Norm} & {F1}     \\
                  \midrule
EXIST2023\_gold    & 6.2431   & 1               & 0.8926          & 1.6007   & 1               & 1      \\
EXIST2023\_majority\_class & -5.6674  & 0.6592          & 4.9745          & -1.0391  & 0.4185          & 0.1545 \\
EXIST2023\_minority\_class & -28.7093 & 0               & 8.757           & -2.939   & 0               & 0.0332 \\
\midrule
AI-UPV\_1  (Model Selection)                    & -4.013   & 0.7066          & 3.0365          & 0.1273   & 0.6754          & \B 0.5123 \\
AI-UPV\_2  (Ensemble)                      & \B -1.6145  &  \B 0.7752          & \B 2.1641          & \B 0.1827   & \B 0.6876          & \B 0.5124 \\
AI-UPV\_3  (Adj. Ensemble)                 & -1.6568  & 0.774           & 2.5311          & 0.1618   & 0.683           &  
\B 0.5142 \\
                               \bottomrule\\
    \end{tabular}%
}

\end{table}

\begin{table}[h]
\caption{Results of the proposed approaches for Task 2 on EN (English) instances.}
\label{tab:res-task2-EN}
\resizebox{\textwidth}{!}{%
\sisetup{detect-weight=true,mode=text,
  round-mode = places,      
  round-precision = 2,      
}
\begin{tabular}{lSSSSSS}
                               \toprule
                               & {ICM-Soft} & {ICM-Soft   Norm} & {Cross   Entropy} & {ICM-Hard} & {ICM-Hard   Norm} & {F1}     \\
                               \midrule
EXIST2023\_gold    & 6.1178   & 1               & 0.9354          & 1.4449   & 1               & 1      \\
EXIST2023\_majority\_class & -5.2028  & 0.7518          & 4.2291          & -0.8529  & 0.5327          & 0.1667 \\
EXIST2023\_minority\_class & -39.4948 & 0               & 8.9579          & -3.4728  & 0               & 0.022  \\
\midrule
AI-UPV\_1  (Model Selection)                    & -5.0889  & 0.7543          & 2.9956          & 0.0876   & 0.724           & 0.4547 \\
AI-UPV\_2  (Ensemble)                      &  \B -1.9067  & \B 0.8241          & \B 2.1761          & \B 0.1942   & \B 0.7457          & \B 0.4758 \\
AI-UPV\_3  (Adj. Ensemble)                 & -2.0569  & 0.8208          & 2.555           & 0.1858   & 0.744           & 0.4718\\\bottomrule
\end{tabular}%
}

\end{table}
\noindent It is important to note that, although  Task 2 is more challenging than task 1, the proposed approach performs in a very competitive way when comparing the obtained ICM-Soft-Norm and ICM-Hard Norm with the corresponding values obtained in Task 1 (See Table \ref{tab:res-task1-ALL}). This suggests that by deepening the labeling taxonomy, the model is prone to make suitable predictions at lower granularity  levels. This means that the intention of sexism can be considered as simpler task than distinguishing what is sexist from what is not. These results are also confirmed by the performance on single languages, where no substantial differences are observed between English and Spanish (Tables \ref{tab:res-task2-EN} and \ref{tab:res-task2-ES} respectively).

\begin{table}[h]
\caption{Results of the proposed approaches for Task 3 on all test instances (English + Spanish).}
\label{tab:res-task3-all}
\resizebox{\textwidth}{!}{%
\sisetup{detect-weight=true,mode=text,
  round-mode = places,      
  round-precision = 2,      
}
\begin{tabular}{lSSSSS}
\toprule
                            & {ICM-Soft} & {ICM-Soft   Norm} & {ICM-Hard} & {ICM-Hard   Norm} & {F1}     \\
                            \midrule
EXIST2023\_gold\      & 9.4686   & 1               & 2.1533   & 1               & 1      \\
EXIST2023\_majority\_class & -8.7089  & 0.6729          & -15.984  & 0.2898          & 0.1069 \\
EXIST2023\_minority\_class & -46.108  & 0               & -31.295  & 0               & 0.0288 \\
\midrule
AI-UPV\_1  (Model Selection)                    & -3.3437  & 0.7695          & \B -0.1862  & \B 0.5571          & 0.4732 \\
AI-UPV\_2  (Ensemble)                      & -2.5616  & 0.7835          & -0.2516  & 0.5448          & \B 0.4757 \\
AI-UPV\_3  (Adj. Ensemble)                 & \B -2.3183  & \B 0.7879          & -0.5788  & 0.4828          & 0.4195\\
\bottomrule
\end{tabular}%
}

\end{table}
\noindent Table \ref{tab:res-task3-all} finally shows the results for the EXIST Task 3 on the  entire dataset, where the main goal was to distinguish between different types of sexism. In this case, the best-performing model, i.e. AI-UPV\_3, ranked first in the global ranking considering the soft evaluation, achieving an ICM-Soft Norm value equal to 0.79. Differently from what has been observed for the previous tasks, the prediction adjustment significantly improves the ensemble performances related to the soft evaluation.  In this case, the task is even more complex than the previous ones because it is based on a fine-grained labeling schema that also depends on the previous levels of annotation. By comparing the current ICM-Soft value achieved by the AU-UPV\_3 approach (0.79) with the corresponding one reported in Table \ref{tab:res-task2-all} (0.80), the capabilities of the model at the lower level of the hierarchy are still maintained. This confirms the ability of the model at addressing fine-grained predictions. Analogously results can be grasped by looking at Tables \ref{tab:res-task3-ES} and \ref{tab:res-task3-EN} that focus on Spanish and English instances, respectively.

\begin{table}[h]
\caption{Results of the proposed approaches for Task 3 on ES (Spanish) instances.}
\label{tab:res-task3-ES}
\resizebox{\textwidth}{!}{%
\sisetup{detect-weight=true,mode=text,
  round-mode = places,      
  round-precision = 2,      
}
\begin{tabular}{lSSSSS}
\toprule
                            & {ICM-Soft} & {ICM-Soft   Norm} & {ICM-Hard} & {ICM-Hard   Norm} & {F1}     \\
                            \midrule
EXIST2023\_gold      & 9.6071   & 1               & 2.2393   & 1               & 1      \\
EXIST2023\_majority\_class & -9.0314  & 0.6613          & -1.7269  & 0.2865          & 0.103  \\
EXIST2023\_minority\_class & -45.426  & 0               & -3.3196  & 0               & 0.0276 \\
\midrule
AI-UPV\_1  (Model Selection)                    & -3.6482  & 0.7591          & \B -0.2284  &\B 0.5561          & \B 0.4821 \\
AI-UPV\_2  (Ensemble)                      & -2.7417  & \B 0.7756          & -0.3109  & 0.5412          & \B 0.4822 \\
AI-UPV\_3  (Adj. Ensemble)                 & \B -2.5779  &\B  0.7786          & -0.6549  & 0.4794          & 0.4218\\
\bottomrule
\end{tabular}%
}

\end{table}

\begin{table}[h]
\caption{Results of the proposed approaches for Task 3 on EN (English) instances.}
\label{tab:res-task3-EN}
\resizebox{\textwidth}{!}{%
\sisetup{detect-weight=true,mode=text,
  round-mode = places,      
  round-precision = 2,      
}
\begin{tabular}{lSSSSS}
                               \toprule
Run                            & {ICM-Soft} & {ICM-Soft   Norm} & {ICM-Hard} & {ICM-Hard   Norm} & {F1}     \\

\midrule
EXIST2023\_test\_gold\_soft      & 9.1255   & 1               & 2.0402   & 1               & 1      \\
EXIST2023\_test\_majority\_class & -8.2105  & 0.6908          & -1.4563  & 0.2962          & 0.1111 \\
EXIST2023\_test\_minority\_class & -46.9473 & 0               & -2.9279  & 0               & 0.0301 \\
\midrule
AI-UPV\_1  (Model Selection)                    & -3.0176  & 0.7834          & \B -0.1528  & \B 0.5586          & \B 0.4571 \\
AI-UPV\_2  (Ensemble)                      & -2.3916  & 0.7946          & -0.1958  & 0.5499          & \B 0.4601 \\
AI-UPV\_3  (Adj. Ensemble)                 & \B -2.0487  & \B 0.8007          & -0.5023  & 0.4882          & 0.4122\\
\bottomrule
\end{tabular}%
}

\end{table}



\section{Conclusion}\label{sec:conclusion}
In this work, we proposed a large language models-based ensemble strategy to address the task of sexism identification under the paradigm of learning with disagreements.\\
The achieved results highlight how the adoption of ensembles can significantly improve the results obtained by the individual models. We developed a simple average ensemble strategy, however, the investigation of more complex ensemble strategies with the proposed system could be conducted in future works. 
Additionally, the proposed ensemble model can easily be extended to include also other Large Language Models that have shown promising performance on similar tasks (e.g., LlaMA \cite{touvron2023llama} or ELECTRA \cite{clark2020electra}).\\
Another possible improvement of the proposed approach refers to the inclusion of additional features. While the proposed approach only considers the text within the tweet, sentiment information and lexical characteristics (e.g., the usage of uppercase or emoji) have been shown to be important clues for hate-related tasks. 
Moreover, despite previous works showing the importance of author profiling and demonstrating the utility of exploiting annotators' characteristics in disagreement detection, this information for the test dataset has not yet been released -- making those strategies unfeasible for the participation in the challenge -- but is a line of work that we plan to investigate in the near future.

\begin{acknowledgments}
This research is partially supported by the ARC Centre of Excellence for Automated Decision-Making and Society (ADM+S), funded by the Australian Research Council (CE200100005). Damiano Spina is the recipient of an ARC DECRA Research Fellowship (DE200100064). \\
Angel Felipe Magnoss\~{a}o de Paula has received a mobility grant for doctoral 
students by the Universitat Polit\`{e}cnica de Val\`{e}ncia to visit RMIT University.\\
The work of Elisabetta Fersini has been partially funded by the European Union -– NextGenerationEU under the National Research Centre For HPC, Big Data and Quantum Computing -- Spoke 9 -- Digital Society and Smart Cities (PNRR-MUR), and by REGAINS -- Excellence Department Research Project. 
   
\end{acknowledgments}


\bibliography{main}

\begin{thebibliography}{40}
\expandafter\ifx\csname natexlab\endcsname\relax\def\natexlab#1{#1}\fi
\providecommand{\url}[1]{\texttt{#1}}
\providecommand{\href}[2]{#2}
\providecommand{\path}[1]{#1}
\providecommand{\DOIprefix}{doi:}
\providecommand{\ArXivprefix}{arXiv:}
\providecommand{\URLprefix}{URL: }
\providecommand{\Pubmedprefix}{pmid:}
\providecommand{\doi}[1]{\href{http://dx.doi.org/#1}{\path{#1}}}
\providecommand{\Pubmed}[1]{\href{pmid:#1}{\path{#1}}}
\providecommand{\bibinfo}[2]{#2}
\ifx\xfnm\relax \def\xfnm[#1]{\unskip,\space#1}\fi
\bibitem[{Plaza et~al.(2023{\natexlab{a}})Plaza, {Carrillo-de-Albornoz},
  Morante, Amigó, Gonzalo, Spina, and Rosso}]{plaza2023overview}
\bibinfo{author}{L.~Plaza}, \bibinfo{author}{J.~{Carrillo-de-Albornoz}},
  \bibinfo{author}{R.~Morante}, \bibinfo{author}{E.~Amigó},
  \bibinfo{author}{J.~Gonzalo}, \bibinfo{author}{D.~Spina},
  \bibinfo{author}{P.~Rosso},
\newblock \bibinfo{title}{{Overview of {EXIST} 2023 -- Learning with
  Disagreement for Sexism Identification and Characterization}},
\newblock in: \bibinfo{editor}{A.~Arampatzis}, \bibinfo{editor}{E.~Kanoulas},
  \bibinfo{editor}{T.~Tsikrika}, \bibinfo{editor}{S.~Vrochidis},
  \bibinfo{editor}{A.~Giachanou}, \bibinfo{editor}{D.~Li},
  \bibinfo{editor}{M.~Aliannejadi}, \bibinfo{editor}{M.~Vlachos},
  \bibinfo{editor}{G.~Faggioli}, \bibinfo{editor}{N.~Ferro} (Eds.),
  \bibinfo{booktitle}{Experimental IR Meets Multilinguality, Multimodality, and
  Interaction}, \bibinfo{address}{Thessaloniki, Greece},
  \bibinfo{year}{2023}{\natexlab{a}}.
\bibitem[{Plaza et~al.(2023{\natexlab{b}})Plaza, {Carrillo-de-Albornoz},
  Morante, Amigó, Gonzalo, Spina, and Rosso}]{plaza2023extended}
\bibinfo{author}{L.~Plaza}, \bibinfo{author}{J.~{Carrillo-de-Albornoz}},
  \bibinfo{author}{R.~Morante}, \bibinfo{author}{E.~Amigó},
  \bibinfo{author}{J.~Gonzalo}, \bibinfo{author}{D.~Spina},
  \bibinfo{author}{P.~Rosso},
\newblock \bibinfo{title}{{Overview of {EXIST} 2023 -- Learning with
  Disagreement for Sexism Identification and Characterization (Extended
  Overview)}},
\newblock in: \bibinfo{editor}{M.~Aliannejadi}, \bibinfo{editor}{G.~Faggioli},
  \bibinfo{editor}{N.~Ferro}, \bibinfo{editor}{M.~Vlachos} (Eds.),
  \bibinfo{booktitle}{Working Notes of {CLEF} 2023 -- Conference and Labs of
  the Evaluation Forum}, \bibinfo{year}{2023}{\natexlab{b}}.
\bibitem[{{Baris Schlicht} et~al.(2021{\natexlab{a}}){Baris Schlicht},
  {Magnoss{\~a}o de Paula}, and {Rosso}}]{schlicht2021upv}
\bibinfo{author}{I.~{Baris Schlicht}}, \bibinfo{author}{A.~F. {Magnoss{\~a}o de
  Paula}}, \bibinfo{author}{P.~{Rosso}},
\newblock \bibinfo{title}{{UPV at CheckThat! 2021: Mitigating Cultural
  Differences for Identifying Multilingual Check-worthy Claims}},
\newblock in: \bibinfo{booktitle}{Proceedings of The 12th Conference and Labs
  of the Evaluation Forum (CLEF)}, volume \bibinfo{volume}{2936},
  \bibinfo{year}{2021}{\natexlab{a}}, pp. \bibinfo{pages}{465--475}.
\bibitem[{{Baris Schlicht} et~al.(2021{\natexlab{b}}){Baris Schlicht},
  {Magnoss{\~a}o de Paula}, and {Rosso}}]{Ipek2021TREC}
\bibinfo{author}{I.~{Baris Schlicht}}, \bibinfo{author}{A.~F. {Magnoss{\~a}o de
  Paula}}, \bibinfo{author}{P.~{Rosso}},
\newblock \bibinfo{title}{{UPV at TREC Health Misinformation Track 2021 Ranking
  with SBERT and Quality Estimators}},
\newblock in: \bibinfo{booktitle}{Proceedings of The Thirtieth Text REtrieval
  Conference (TREC)}, \bibinfo{year}{2021}{\natexlab{b}}.
\bibitem[{Baris~Schlicht and Magnossão~de Paula(2021)}]{schlicht2021unified}
\bibinfo{author}{I.~Baris~Schlicht}, \bibinfo{author}{A.~F. Magnossão~de
  Paula},
\newblock \bibinfo{title}{{Unified and Multilingual Author Profiling for
  Detecting Haters}},
\newblock in: \bibinfo{booktitle}{Proceedings of The 12th Conference and Labs
  of the Evaluation Forum (CLEF)}, volume \bibinfo{volume}{2936},
  \bibinfo{year}{2021}, pp. \bibinfo{pages}{1837--1845}.
\bibitem[{Samory et~al.(2021)Samory, Sen, Kohne, Fl{\"o}ck, and
  Wagner}]{samory2021call}
\bibinfo{author}{M.~Samory}, \bibinfo{author}{I.~Sen},
  \bibinfo{author}{J.~Kohne}, \bibinfo{author}{F.~Fl{\"o}ck},
  \bibinfo{author}{C.~Wagner},
\newblock \bibinfo{title}{{“Call me sexist, but...”: Revisiting Sexism
  Detection Using Psychological Scales and Adversarial Samples}},
\newblock in: \bibinfo{booktitle}{Proceedings of the International AAAI
  Conference on Web and Social Media}, volume~\bibinfo{volume}{15},
  \bibinfo{year}{2021}, pp. \bibinfo{pages}{573--584}.
\bibitem[{Magnoss{\~a}o~de Paula et~al.(2021)Magnoss{\~a}o~de Paula, da~Silva,
  and Schlicht}]{de2021sexism}
\bibinfo{author}{A.~F. Magnoss{\~a}o~de Paula}, \bibinfo{author}{R.~F.
  da~Silva}, \bibinfo{author}{I.~B. Schlicht},
\newblock \bibinfo{title}{{Sexism Prediction in Spanish and English Tweets
  Using Monolingual and Multilingual BERT and Ensemble Models}},
\newblock in: \bibinfo{booktitle}{Proceedings of the Iberian Languages
  Evaluation Forum (IberLEF 2021) co-located with the XXXVII International
  Conference of the Spanish Society for Natural Language Processing (SEPLN)},
  volume \bibinfo{volume}{2943}, \bibinfo{year}{2021}, pp.
  \bibinfo{pages}{356--373}.
\bibitem[{Magnoss{\~a}o~de Paula and da~Silva(2022)}]{de2022detection}
\bibinfo{author}{A.~F. Magnoss{\~a}o~de Paula}, \bibinfo{author}{R.~F.
  da~Silva},
\newblock \bibinfo{title}{{Detection and Classification of Sexism on Social
  Media Using Multiple Languages, Transformers, and Ensemble Models}},
\newblock in: \bibinfo{booktitle}{Proceedings of the Iberian Languages
  Evaluation Forum (IberLEF 2022) co-located with the XXXVIII International
  Conference of the Spanish Society for Natural Language Processing (SEPLN)},
  volume \bibinfo{volume}{3202}, \bibinfo{year}{2022}.
\bibitem[{Campisi et~al.(2018)Campisi, Corchs, Fersini, Gasparini, and
  Mantovani}]{campisi2018automatic}
\bibinfo{author}{G.~Campisi}, \bibinfo{author}{S.~Corchs},
  \bibinfo{author}{E.~Fersini}, \bibinfo{author}{F.~Gasparini},
  \bibinfo{author}{M.~Mantovani},
\newblock \bibinfo{title}{Automatic detection of sexist content in memes},
\newblock \bibinfo{journal}{Image} \bibinfo{volume}{46} (\bibinfo{year}{2018})
  \bibinfo{pages}{53--9}.
\bibitem[{Fersini et~al.(2019)Fersini, Gasparini, and
  Corchs}]{fersini2019detecting}
\bibinfo{author}{E.~Fersini}, \bibinfo{author}{F.~Gasparini},
  \bibinfo{author}{S.~Corchs},
\newblock \bibinfo{title}{Detecting sexist meme on the web: A study on textual
  and visual cues},
\newblock in: \bibinfo{booktitle}{2019 8th International Conference on
  Affective Computing and Intelligent Interaction Workshops and Demos (ACIIW)},
  \bibinfo{organization}{IEEE}, \bibinfo{year}{2019}, pp.
  \bibinfo{pages}{226--231}.
\bibitem[{Rodr{\'\i}guez-S{\'a}nchez et~al.(2021)Rodr{\'\i}guez-S{\'a}nchez,
  Carrillo-de Albornoz, Plaza, Gonzalo, Rosso, Comet, and
  Donoso}]{rodriguez2021overview}
\bibinfo{author}{F.~Rodr{\'\i}guez-S{\'a}nchez},
  \bibinfo{author}{J.~Carrillo-de Albornoz}, \bibinfo{author}{L.~Plaza},
  \bibinfo{author}{J.~Gonzalo}, \bibinfo{author}{P.~Rosso},
  \bibinfo{author}{M.~Comet}, \bibinfo{author}{T.~Donoso},
\newblock \bibinfo{title}{{Overview of EXIST 2021: sEXism Identification in
  Social neTworks}},
\newblock \bibinfo{journal}{Procesamiento del Lenguaje Natural}
  \bibinfo{volume}{67} (\bibinfo{year}{2021}) \bibinfo{pages}{195--207}.
\bibitem[{Rodr{\'\i}guez-S{\'a}nchez et~al.(2022)Rodr{\'\i}guez-S{\'a}nchez,
  Carrillo-de Albornoz, Plaza, Mendieta-Arag{\'o}n, Marco-Rem{\'o}n, Makeienko,
  Plaza, Gonzalo, Spina, and Rosso}]{rodriguez2022overview}
\bibinfo{author}{F.~Rodr{\'\i}guez-S{\'a}nchez},
  \bibinfo{author}{J.~Carrillo-de Albornoz}, \bibinfo{author}{L.~Plaza},
  \bibinfo{author}{A.~Mendieta-Arag{\'o}n},
  \bibinfo{author}{G.~Marco-Rem{\'o}n}, \bibinfo{author}{M.~Makeienko},
  \bibinfo{author}{M.~Plaza}, \bibinfo{author}{J.~Gonzalo},
  \bibinfo{author}{D.~Spina}, \bibinfo{author}{P.~Rosso},
\newblock \bibinfo{title}{{Overview of EXIST 2022: sEXism Identification in
  Social neTworks}},
\newblock \bibinfo{journal}{Procesamiento del Lenguaje Natural}
  \bibinfo{volume}{69} (\bibinfo{year}{2022}) \bibinfo{pages}{229--240}.
\bibitem[{Kirk et~al.(2023)Kirk, Yin, Vidgen, and
  R{\"o}ttger}]{kirk2023semeval}
\bibinfo{author}{H.~R. Kirk}, \bibinfo{author}{W.~Yin},
  \bibinfo{author}{B.~Vidgen}, \bibinfo{author}{P.~R{\"o}ttger},
\newblock \bibinfo{title}{{SemEval-2023 Task 10: Explainable Detection of
  Online Sexism}},
\newblock \bibinfo{journal}{arXiv preprint arXiv:2303.04222}
  (\bibinfo{year}{2023}).
\bibitem[{Rydelek et~al.(2023)Rydelek, Dementieva, and Groh}]{rydelek2023adamr}
\bibinfo{author}{A.~Rydelek}, \bibinfo{author}{D.~Dementieva},
  \bibinfo{author}{G.~Groh},
\newblock \bibinfo{title}{{AdamR at SemEval-2023 Task 10: Solving the Class
  Imbalance Problem in Sexism Detection with Ensemble Learning}},
\newblock \bibinfo{journal}{arXiv preprint arXiv:2305.08636}
  (\bibinfo{year}{2023}).
\bibitem[{Rallabandi et~al.(2023)Rallabandi, Singhal, and
  Seth}]{rallabandi2023sss}
\bibinfo{author}{S.~Rallabandi}, \bibinfo{author}{S.~Singhal},
  \bibinfo{author}{P.~Seth},
\newblock \bibinfo{title}{{SSS at SemEval-2023 Task 10: Explainable Detection
  of Online Sexism using Majority Voted Fine-Tuned Transformers}},
\newblock \bibinfo{journal}{arXiv preprint arXiv:2304.03518}
  (\bibinfo{year}{2023}).
\bibitem[{Magnossão~de Paula and Baris~Schlicht(2021)}]{magnossao2021ai}
\bibinfo{author}{A.~F. Magnossão~de Paula},
  \bibinfo{author}{I.~Baris~Schlicht},
\newblock \bibinfo{title}{{AI-UPV at IberLEF-2021 DETOXIS task: Toxicity
  Detection in Immigration-Related Web News Comments Using Transformers and
  Statistical Models}},
\newblock in: \bibinfo{booktitle}{Proceedings of the Iberian Languages
  Evaluation Forum (IberLEF 2021) co-located with the XXXVII International
  Conference of the Spanish Society for Natural Language Processing (SEPLN)},
  \bibinfo{year}{2021}, pp. \bibinfo{pages}{547--566}.
\bibitem[{Magnossão~de Paula et~al.(2022)Magnossão~de Paula, Rosso, Bensalem,
  and Zaghouani}]{de2022upv}
\bibinfo{author}{A.~F. Magnossão~de Paula}, \bibinfo{author}{P.~Rosso},
  \bibinfo{author}{I.~Bensalem}, \bibinfo{author}{W.~Zaghouani},
\newblock \bibinfo{title}{{UPV at the Arabic Hate Speech 2022 Shared Task:
  Offensive Language and Hate Speech Detection Using Transformers and Ensemble
  Models}},
\newblock in: \bibinfo{booktitle}{Proceedings of the 5th Workshop on
  Open-Source Arabic Corpora and Processing Tools with Shared Tasks on Qur'an
  QA and Fine-Grained Hate Speech Detection}, \bibinfo{year}{2022}, pp.
  \bibinfo{pages}{181--185}.
\bibitem[{Magnossão~de Paula et~al.(2023)Magnossão~de Paula, Bensalem, Rosso,
  and Zaghouani}]{depaula2023transformers}
\bibinfo{author}{A.~F. Magnossão~de Paula}, \bibinfo{author}{I.~Bensalem},
  \bibinfo{author}{P.~Rosso}, \bibinfo{author}{W.~Zaghouani},
\newblock \bibinfo{title}{{Transformers and Ensemble Methods: A Solution for
  Hate Speech Detection in Arabic Languages}},
\newblock in: \bibinfo{booktitle}{Proceedings of the 1st Natural Language
  Processing (NLP) Challenge at Centre De Recherche Sur L'information
  Scientifique et Technique (CERIST)}, \bibinfo{year}{2023}.
\bibitem[{Fersini et~al.(2018)Fersini, Rosso, and
  Anzovino}]{fersini2018overview}
\bibinfo{author}{E.~Fersini}, \bibinfo{author}{P.~Rosso},
  \bibinfo{author}{M.~Anzovino},
\newblock \bibinfo{title}{{Overview of the Task on Automatic Misogyny
  Identification at IberEval 2018}},
\newblock \bibinfo{journal}{IberEval @ SEPLN} \bibinfo{volume}{2150}
  (\bibinfo{year}{2018}) \bibinfo{pages}{214--228}.
\bibitem[{Gandhi et~al.(2020)Gandhi, Kokkula, Chaudhuri, Magnani, Stanley,
  Ahmadi, Kandaswamy, Ovenc, and Mannor}]{gandhi2020scalable}
\bibinfo{author}{S.~Gandhi}, \bibinfo{author}{S.~Kokkula},
  \bibinfo{author}{A.~Chaudhuri}, \bibinfo{author}{A.~Magnani},
  \bibinfo{author}{T.~Stanley}, \bibinfo{author}{B.~Ahmadi},
  \bibinfo{author}{V.~Kandaswamy}, \bibinfo{author}{O.~Ovenc},
  \bibinfo{author}{S.~Mannor},
\newblock \bibinfo{title}{{Scalable Detection of Offensive and Non-compliant
  Content / Logo in Product Images}},
\newblock in: \bibinfo{booktitle}{Proceedings of the IEEE/CVF Winter Conference
  on Applications of Computer Vision}, \bibinfo{year}{2020}, pp.
  \bibinfo{pages}{2247--2256}.
\bibitem[{Tabone et~al.(2021)Tabone, Camilleri, Bonnici, Cristina, Farrugia,
  and Borg}]{tabone2021pornographic}
\bibinfo{author}{A.~Tabone}, \bibinfo{author}{K.~Camilleri},
  \bibinfo{author}{A.~Bonnici}, \bibinfo{author}{S.~Cristina},
  \bibinfo{author}{R.~Farrugia}, \bibinfo{author}{M.~Borg},
\newblock \bibinfo{title}{{Pornographic Content Classification Using
  Deep-Learning}},
\newblock in: \bibinfo{booktitle}{21st ACM Symposium on Document Engineering},
  \bibinfo{year}{2021}, pp. \bibinfo{pages}{1--10}.
\bibitem[{Hor et~al.(2021)Hor, Karim, Abdullah, AlDahoul, Mansor, Fauzi, See,
  and Wazir}]{SoAPornographyDetection}
\bibinfo{author}{S.~L. Hor}, \bibinfo{author}{H.~A. Karim},
  \bibinfo{author}{M.~H.~L. Abdullah}, \bibinfo{author}{N.~AlDahoul},
  \bibinfo{author}{S.~Mansor}, \bibinfo{author}{M.~F.~A. Fauzi},
  \bibinfo{author}{J.~See}, \bibinfo{author}{A.~S.~B. Wazir},
\newblock \bibinfo{title}{{An Evaluation of State-of-the-Art Object Detectors
  for Pornography Detection}},
\newblock in: \bibinfo{booktitle}{IEEE International Conference on Signal and
  Image Processing Applications (ICSIPA)}, \bibinfo{year}{2021}, pp.
  \bibinfo{pages}{191--196}.
\bibitem[{Kiela et~al.(2020)Kiela, Firooz, Mohan, Goswami, Singh, Ringshia, and
  Testuggine}]{kiela2020hateful}
\bibinfo{author}{D.~Kiela}, \bibinfo{author}{H.~Firooz},
  \bibinfo{author}{A.~Mohan}, \bibinfo{author}{V.~Goswami},
  \bibinfo{author}{A.~Singh}, \bibinfo{author}{P.~Ringshia},
  \bibinfo{author}{D.~Testuggine},
\newblock \bibinfo{title}{{The Hateful Memes Challenge: Detecting Hate Speech
  in Multimodal Memes}},
\newblock \bibinfo{journal}{Advances in Neural Information Processing Systems}
  \bibinfo{volume}{33} (\bibinfo{year}{2020}) \bibinfo{pages}{2611--2624}.
\bibitem[{Fersini et~al.(2022)Fersini, Gasparini, Rizzi, Saibene, Chulvi,
  Rosso, Lees, and Sorensen}]{fersini2022semeval}
\bibinfo{author}{E.~Fersini}, \bibinfo{author}{F.~Gasparini},
  \bibinfo{author}{G.~Rizzi}, \bibinfo{author}{A.~Saibene},
  \bibinfo{author}{B.~Chulvi}, \bibinfo{author}{P.~Rosso},
  \bibinfo{author}{A.~Lees}, \bibinfo{author}{J.~Sorensen},
\newblock \bibinfo{title}{{SemEval-2022 Task 5: Multimedia Automatic Misogyny
  Identification}},
\newblock in: \bibinfo{booktitle}{Proceedings of the 16th International
  Workshop on Semantic Evaluation (SemEval-2022)}, \bibinfo{year}{2022}, pp.
  \bibinfo{pages}{533--549}.
\bibitem[{Alba-Cepero et~al.(2022)Alba-Cepero, Cabezas-Puerto,
  L{\'o}pez-Batista, and Moreno-Montero}]{alba2022bias}
\bibinfo{author}{J.~Alba-Cepero}, \bibinfo{author}{M.~Cabezas-Puerto},
  \bibinfo{author}{V.~F. L{\'o}pez-Batista}, \bibinfo{author}{{\'A}.~M.
  Moreno-Montero},
\newblock \bibinfo{title}{{Bias Analysis on Twitter}},
\newblock in: \bibinfo{booktitle}{New Trends in Disruptive Technologies, Tech
  Ethics and Artificial Intelligence: The DITTET 2022 Collection},
  \bibinfo{publisher}{Springer}, \bibinfo{year}{2022}, pp.
  \bibinfo{pages}{131--142}.
\bibitem[{Wiegand et~al.(2019)Wiegand, Ruppenhofer, and
  Kleinbauer}]{wiegand2019detection}
\bibinfo{author}{M.~Wiegand}, \bibinfo{author}{J.~Ruppenhofer},
  \bibinfo{author}{T.~Kleinbauer},
\newblock \bibinfo{title}{{D}etection of {A}busive {L}anguage: the {P}roblem of
  {B}iased {D}atasets},
\newblock in: \bibinfo{booktitle}{Proceedings of the 2019 Conference of the
  North {A}merican Chapter of the Association for Computational Linguistics:
  Human Language Technologies, Volume 1 (Long and Short Papers)},
  \bibinfo{publisher}{Association for Computational Linguistics},
  \bibinfo{address}{Minneapolis, Minnesota}, \bibinfo{year}{2019}, pp.
  \bibinfo{pages}{602--608}. \DOIprefix\doi{10.18653/v1/N19-1060}.
\bibitem[{Zhou(2021)}]{zhou2021challenges}
\bibinfo{author}{X.~Zhou}, \bibinfo{title}{{Challenges in Automated Debiasing
  for Toxic Language Detection}}, \bibinfo{publisher}{University of
  Washington}, \bibinfo{year}{2021}.
\bibitem[{Uma et~al.(2021)Uma, Fornaciari, Hovy, Paun, Plank, and
  Poesio}]{uma2021learning}
\bibinfo{author}{A.~N. Uma}, \bibinfo{author}{T.~Fornaciari},
  \bibinfo{author}{D.~Hovy}, \bibinfo{author}{S.~Paun},
  \bibinfo{author}{B.~Plank}, \bibinfo{author}{M.~Poesio},
\newblock \bibinfo{title}{{Learning from Disagreement: A Survey}},
\newblock \bibinfo{journal}{Journal of Artificial Intelligence Research}
  \bibinfo{volume}{72} (\bibinfo{year}{2021}) \bibinfo{pages}{1385--1470}.
  \DOIprefix\doi{10.1613/jair.1.12752}.
\bibitem[{Kalra and Zubiaga(2021)}]{kalra2021sexism}
\bibinfo{author}{A.~Kalra}, \bibinfo{author}{A.~Zubiaga},
\newblock \bibinfo{title}{{Sexism Identification in Tweets and Gabs using Deep
  Neural Networks}},
\newblock \bibinfo{journal}{arXiv preprint arXiv:2111.03612}
  (\bibinfo{year}{2021}).
\bibitem[{Amig{\'o} and Delgado(2022)}]{amigo2022evaluating}
\bibinfo{author}{E.~Amig{\'o}}, \bibinfo{author}{A.~Delgado},
\newblock \bibinfo{title}{{Evaluating Extreme Hierarchical Multi-label
  Classification}},
\newblock in: \bibinfo{booktitle}{Proceedings of the 60th Annual Meeting of the
  Association for Computational Linguistics (Volume 1: Long Papers)},
  \bibinfo{publisher}{Association for Computational Linguistics},
  \bibinfo{address}{Dublin, Ireland}, \bibinfo{year}{2022}, pp.
  \bibinfo{pages}{5809--5819}. \DOIprefix\doi{10.18653/v1/2022.acl-long.399}.
\bibitem[{Devlin et~al.(2019)Devlin, Chang, Lee, and
  Toutanova}]{devlin2019bert}
\bibinfo{author}{J.~Devlin}, \bibinfo{author}{M.-W. Chang},
  \bibinfo{author}{K.~Lee}, \bibinfo{author}{K.~Toutanova},
\newblock \bibinfo{title}{{{BERT}: Pre-training of Deep Bidirectional
  Transformers for Language Understanding}},
\newblock in: \bibinfo{booktitle}{Proceedings of the 2019 Conference of the
  North {A}merican Chapter of the Association for Computational Linguistics:
  Human Language Technologies, Volume 1 (Long and Short Papers)},
  \bibinfo{publisher}{Association for Computational Linguistics},
  \bibinfo{address}{Minneapolis, Minnesota}, \bibinfo{year}{2019}, pp.
  \bibinfo{pages}{4171--4186}. \DOIprefix\doi{10.18653/v1/N19-1423}.
\bibitem[{Sch{\"u}tz et~al.(2021)Sch{\"u}tz, Boeck, Liakhovets,
  Slijep{\v{c}}evi{\'c}, Kirchknopf, Hecht, Bogensperger, Schlarb, Schindler,
  and Zeppelzauer}]{schutz2021automatic}
\bibinfo{author}{M.~Sch{\"u}tz}, \bibinfo{author}{J.~Boeck},
  \bibinfo{author}{D.~Liakhovets}, \bibinfo{author}{D.~Slijep{\v{c}}evi{\'c}},
  \bibinfo{author}{A.~Kirchknopf}, \bibinfo{author}{M.~Hecht},
  \bibinfo{author}{J.~Bogensperger}, \bibinfo{author}{S.~Schlarb},
  \bibinfo{author}{A.~Schindler}, \bibinfo{author}{M.~Zeppelzauer},
\newblock \bibinfo{title}{{Automatic Sexism Detection with Multilingual
  Transformer Models}},
\newblock in: \bibinfo{booktitle}{Proceedings of the Iberian Languages
  Evaluation Forum (IberLEF 2021) co-located with the XXXVII International
  Conference of the Spanish Society for Natural Language Processing (SEPLN)},
  volume \bibinfo{volume}{2943}, \bibinfo{year}{2021}. \URLprefix
  \url{https://ceur-ws.org/Vol-2943/exist_paper1.pdf}.
\bibitem[{Bengoetxea and Aguirregoitia(2022)}]{bengoetxea2022multiaztertest}
\bibinfo{author}{K.~Bengoetxea}, \bibinfo{author}{A.~Aguirregoitia},
\newblock \bibinfo{title}{{Multiaztertest@Exist-Iberlef2022: Sexism
  Identification in Social Networks}},
\newblock in: \bibinfo{booktitle}{Proceedings of the Iberian Languages
  Evaluation Forum (IberLEF 2022) co-located with the XXXVIII International
  Conference of the Spanish Society for Natural Language Processing (SEPLN)},
  volume \bibinfo{volume}{3202}, \bibinfo{year}{2022}. \URLprefix
  \url{https://ceur-ws.org/Vol-3202/exist-paper8.pdf}.
\bibitem[{Conneau et~al.(2019)Conneau, Khandelwal, Goyal, Chaudhary, Wenzek,
  Guzm{\'a}n, Grave, Ott, Zettlemoyer, and Stoyanov}]{conneau2019unsupervised}
\bibinfo{author}{A.~Conneau}, \bibinfo{author}{K.~Khandelwal},
  \bibinfo{author}{N.~Goyal}, \bibinfo{author}{V.~Chaudhary},
  \bibinfo{author}{G.~Wenzek}, \bibinfo{author}{F.~Guzm{\'a}n},
  \bibinfo{author}{E.~Grave}, \bibinfo{author}{M.~Ott},
  \bibinfo{author}{L.~Zettlemoyer}, \bibinfo{author}{V.~Stoyanov},
\newblock \bibinfo{title}{Unsupervised cross-lingual representation learning at
  scale},
\newblock \bibinfo{journal}{arXiv preprint arXiv:1911.02116}
  (\bibinfo{year}{2019}).
\bibitem[{Ali et~al.(2022)Ali, Farooq, Arshad, Shahzad, and Beg}]{ali2022hate}
\bibinfo{author}{R.~Ali}, \bibinfo{author}{U.~Farooq},
  \bibinfo{author}{U.~Arshad}, \bibinfo{author}{W.~Shahzad},
  \bibinfo{author}{M.~O. Beg},
\newblock \bibinfo{title}{{Hate Speech Detection on Twitter Using Transfer
  Learning}},
\newblock \bibinfo{journal}{Computer Speech \& Language} \bibinfo{volume}{74}
  (\bibinfo{year}{2022}) \bibinfo{pages}{101365}.
\bibitem[{Karim et~al.(2022)Karim, Kanti~Dey, Islam, Shajalal1, and
  Chakravarthi}]{rezaul2022multimodal}
\bibinfo{author}{M.~R. Karim}, \bibinfo{author}{S.~Kanti~Dey},
  \bibinfo{author}{T.~Islam}, \bibinfo{author}{M.~Shajalal1},
  \bibinfo{author}{B.~R. Chakravarthi},
\newblock \bibinfo{title}{{Multimodal Hate Speech Detection from Bengali Memes
  and Texts}},
\newblock in: \bibinfo{booktitle}{International conference on Speech \&
  Language Technology for Low-resource Languages (SPELLL)},
  \bibinfo{year}{2022}, pp. \bibinfo{pages}{1--15}.
\bibitem[{Roy et~al.(2022)Roy, Bhawal, and Subalalitha}]{roy2022hate}
\bibinfo{author}{P.~K. Roy}, \bibinfo{author}{S.~Bhawal},
  \bibinfo{author}{C.~N. Subalalitha},
\newblock \bibinfo{title}{{Hate Speech and Offensive Language Detection in
  Dravidian Languages Using Deep Ensemble Framework}},
\newblock \bibinfo{journal}{Computer Speech \& Language} \bibinfo{volume}{75}
  (\bibinfo{year}{2022}) \bibinfo{pages}{101386}.
\bibitem[{Face(2022)}]{huggingface}
\bibinfo{author}{H.~Face}, \bibinfo{title}{{Transformers: State-of-the-art
  Natural Language Processing}},
  \bibinfo{howpublished}{\url{https://huggingface.co/}}, \bibinfo{year}{2022}.
  \bibinfo{note}{Accessed on May 31, 2023}.
\bibitem[{Touvron et~al.(2023)Touvron, Lavril, Izacard, Martinet, Lachaux,
  Lacroix, Rozière, Goyal, Hambro, Azhar, Rodriguez, Joulin, Grave, and
  Lample}]{touvron2023llama}
\bibinfo{author}{H.~Touvron}, \bibinfo{author}{T.~Lavril},
  \bibinfo{author}{G.~Izacard}, \bibinfo{author}{X.~Martinet},
  \bibinfo{author}{M.-A. Lachaux}, \bibinfo{author}{T.~Lacroix},
  \bibinfo{author}{B.~Rozière}, \bibinfo{author}{N.~Goyal},
  \bibinfo{author}{E.~Hambro}, \bibinfo{author}{F.~Azhar},
  \bibinfo{author}{A.~Rodriguez}, \bibinfo{author}{A.~Joulin},
  \bibinfo{author}{E.~Grave}, \bibinfo{author}{G.~Lample},
  \bibinfo{title}{{LLaMA: Open and Efficient Foundation Language Models}},
  \bibinfo{year}{2023}. \href{http://arxiv.org/abs/2302.13971}{{\tt
  arXiv:2302.13971}}.
\bibitem[{Clark et~al.(2020)Clark, Luong, Le, and Manning}]{clark2020electra}
\bibinfo{author}{K.~Clark}, \bibinfo{author}{M.-T. Luong},
  \bibinfo{author}{Q.~V. Le}, \bibinfo{author}{C.~D. Manning},
  \bibinfo{title}{{ELECTRA: Pre-training Text Encoders as Discriminators Rather
  Than Generators}}, \bibinfo{year}{2020}.
  \href{http://arxiv.org/abs/2003.10555}{{\tt arXiv:2003.10555}}.

\end{thebibliography}
\end{document}